\documentclass{article} 
\usepackage{iclr2025_conference,times}

\usepackage{amsmath,amsfonts,bm}

\def\eqref#1{equation~\ref{#1}}

\def\1{\bm{1}}

\DeclareMathAlphabet{\mathsfit}{\encodingdefault}{\sfdefault}{m}{sl}
\SetMathAlphabet{\mathsfit}{bold}{\encodingdefault}{\sfdefault}{bx}{n}

\usepackage{newfloat}
\usepackage{listings}
\usepackage{newfloat}
\usepackage{listings}
\usepackage{moreverb}
\usepackage{mathtools}
\usepackage{amssymb}
\usepackage{graphics}
\usepackage{subfigure}
\usepackage{caption}
\usepackage{extarrows}
\usepackage{color}
\usepackage{framed}
\usepackage{bm}
\usepackage{mathrsfs}
\usepackage{mathabx}
\usepackage{multirow}
\usepackage{longtable}
\usepackage{paralist}
\usepackage{indentfirst}
\usepackage{relsize}
\usepackage{extarrows}
\usepackage{lineno,xcolor}
\usepackage{upgreek}
\usepackage{bm}
\usepackage{mwe}
\usepackage{xcolor}
\usepackage{booktabs}
\usepackage[flushleft]{threeparttable}
\usepackage{caption}
\usepackage{booktabs}
\usepackage{rotating} 
\usepackage{float}
\usepackage{nameref}

\usepackage{hyperref}
\usepackage{url}

\title{Dynamic Universal Approximation Theory Foundation for Parallelism in CV Neural Networks}

\author{Wei Wang \& Qing Li \\
Department of Comp\\
The Hong Kong Polytechnic University\\
\texttt{weiuat.wang@connect.polyu.hk} \\
\And
Qing Li \\
Department of Comp \\
The Hong Kong Polytechnic University \\
\texttt{qing-prof.li@polyu.edu.hk} \\
}

\iclrfinalcopy 
\begin{document}

\maketitle

\begin{abstract}
Neural networks are increasingly evolving towards training large models with big data, a method that has demonstrated superior performance across many tasks. However, this approach introduces an urgent problem: current deep learning models are serial computing, meaning that as the number of network layers increases, so do the training and inference times. This is unacceptable if deep learning models go deeper. Therefore, this paper proposes a deep learning parallelization strategy based on the Dynamic Universal Approximation Theorem (DUAT). From this foundation, we designed a parallel network called Para-Former to test our theory. Unlike traditional serial models, the inference time of Para-Former does not increase with the number of layers, significantly accelerating the inference speed of multi-layer networks. Experimental results validate the effectiveness of this network.
\end{abstract}

\section{Introduction}
\label{sec:intro}

Currently, deep learning is advancing toward larger-scale models and datasets, a trend driven by the successes of models such as ChatGPT \cite{Radford2018ImprovingLU, brown2020language, achiam2023gpt}, Llama \cite{touvron2023llama}, and ViT \cite{Dosovitskiy2020AnII}. These large models exhibit significantly enhanced performance compared to their predecessors. According to DUAT2LLMs \cite{wang2024universalapproximationtheorybasic1} and DUAT2CVs \cite{wang2024universalapproximationtheorybasic}, both residual-based convolutional neural networks (CNNs) and large-scale Transformer-based models are essentially DUAT functions \cite{wang2024universalapproximationtheorybasic1, wang2024universalapproximationtheorybasic}. DUAT, an evolution of the Universal Approximation Theory (UAT) \cite{Cybenko1989ApproximationBS, Hornik1989MultilayerFN}, enables dynamic function approximation based on input data. From the DUAT perspective, larger models inherently possess a greater dynamic fitting capacity \cite{wang2024universalapproximationtheorybasic1, wang2024universalapproximationtheorybasic}. Consequently, the shift toward larger models is undoubtedly the future of deep learning, and large datasets are essential for training these models to ensure their approximations accurately reflect real-world complexity.

However, this trend towards larger models and datasets inevitably leads to increased demands for computational resources and extended training times. Despite advancements in hardware technology, such as GPUs and TPUs, the rise in time costs remains a significant obstacle. The speed of electrical signal transmission is limited, whereas the theoretical scale of networks has no such upper limit. As network layers increase to tens or even hundreds of thousands, inference times can extend significantly, potentially requiring several minutes or hours for a single batch, making training large models on large datasets exceedingly difficult.

Current solutions include model optimization techniques such as quantization \cite{Jacob2017QuantizationAT,Lu2022AHA,Guo2021IntegerOnlyNN,Lin2023AWQAW}, pruning \cite{Ma2023LLMPrunerOT, Xia2023ShearedLA}, and knowledge distillation \cite{Gou2020KnowledgeDA,Gu2021OpenvocabularyOD}. Additionally, specialized accelerators \cite{Jouppi2017IndatacenterPA,Elbtity2024FlexTPUAF} and edge computing \cite{Shi2016EdgeCV,Gale2020SparseGK}, as well as parallel and distributed computing \cite{Shallue2018MeasuringTE,bennun2018demystifyingparalleldistributeddeep,Sergeev2018HorovodFA}, offer potential solutions. However, these methods do not fundamentally address the inference delay caused by the increasing number of serial network layers.

The root cause lies in the serial computation mechanism of existing deep learning models, where the output of one layer serves as the input for the next. This design originated from early computer vision networks, which used convolutional layers with small receptive fields to accommodate the local correlations and spatiotemporal invariance of image data. Achieving a larger receptive field required deeper network layers, as exemplified by architectures like AlexNet \cite{Krizhevsky2012ImageNetCW} and VGG \cite{Simonyan2014VeryDC}. The introduction of ResNet \cite{He2015DeepRL} marked a paradigm shift, as the residual structure could dynamically fit functions based on input \cite{wang2024universalapproximationtheorybasic1, wang2024universalapproximationtheorybasic}, influencing subsequent network designs and establishing the tradition of multi-layer stacking. While this design was manageable with fewer layers, the increasing depth and number of parameters in modern networks have created a critical issue: slow computation efficiency due to the need for sequential layer-by-layer calculations during inference. This computation method inevitably leads to delays, which become more severe as network depth increases, significantly constraining the progress of computer vision and natural language processing technologies.

Therefore, exploring parallelization techniques for deep learning networks—enabling parallel computation across multiple layers—is crucial for accelerating inference, overcoming time bottlenecks, and advancing computer vision and natural language processing technologies towards greater efficiency and speed. To develop such a parallel network, we build on UAT, integrating DUAT \cite{wang2024universalapproximationtheorybasic1, wang2024universalapproximationtheorybasic} to theoretically propose a foundational approach to parallel network design. To validate our theory, we designed a parallel network, Para-Former, based on the existing ViT architecture. Our contributions include:

\begin{itemize}
\item We propose a parallel network approach based on DUAT. First, we outline the direction for parallel network design based on UAT and then identify the issues associated with this design. Next, we present the fundamental principles and requirements for parallel network design based on DUAT.
\item We analyze the characteristics, advantages, and disadvantages of both serial and parallel networks from the perspective of DUAT.
\item We investigate the issues on performance present in deep learning from a data perspective and give the solution.
\end{itemize}

The structure of this paper is as follows: In Section \ref{section:UAT}, we begin with a brief introduction to the general form of UAT (Section \ref{section:Introduction of UAT}). Next, we present a method for designing parallel networks based on the UAT and discuss the prolems associated with such network designs (Section \ref{section:The Problem in Designing parallel Network Based on UAT}). To address these challenges, we introduce DUAT in Section \ref{section:UAT for Parallel Network}, providing an initial overview in Section \ref{section:Introduction of DUAT} before proposing the Para-former model based on DUAT in Section \ref{section:Para-former}. Finally, we analyze the efficiency improvements of this parallel network in Section \ref{section:Speed-up ratio}. In Section \ref{section:Experiments}, we validate the effectiveness of Para-Former through various evaluations: assessing its feasibility in Section \ref{section:The Feasibility of Para-Former}, examining the impact of network depth in Section \ref{section:Model Depth Effect}, and analyzing the influence of the dataset on network performance in Section \ref{section:Dataset Effect}.

\section{UAT to Parallel Network}
\label{section:UAT}
\subsection{Introduction of UAT}
\label{section:Introduction of UAT}
To develop a parallel network, we start with mathematical forms of UAT. Before formally designing the parallelized network, we will first review the generalized mathematical forms of UAT. The UAT was originally introduced by \citeauthor{Cybenko1989ApproximationBS}, with the following basic form:

\begin{equation}
\setlength{\abovedisplayskip}{1pt}
\setlength{\belowdisplayskip}{1pt}
\begin{aligned}
G(\mathbf{x})=\sum_{j=1}^N \alpha_j \sigma\left(\mathbf{W}_j^{\mathrm{T}} \mathbf{x}+\theta_j\right)
\end{aligned}
\label{Eq:UAP}
\end{equation}
is dense in $C\left(\mathbf{I}_n\right)$. Here, $\mathbf{W}_j \in \mathbb{R}^n$ and $\alpha_j, \theta \in \mathbb{R}$ are parameters $\sigma$ is a sigmoid function. For any $f \in C\left(\mathbf{I}_n\right)$ and $\varepsilon>0$, there exists a function $G(\mathbf{x})$:

\begin{equation}
\begin{aligned}
|G(\mathbf{x})-f(\mathbf{x})|<\varepsilon \quad \text { for all } \quad \mathbf{x} \in \mathbf{I}_n .
\end{aligned}
\label{eq:UPA bound}
\end{equation}

This indicates that when $N$ is sufficiently large, a neural network can approximate any continuous function on a closed interval. \citeauthor{Hornik1989MultilayerFN} further proved that this function can approximate any Borel measurable function. Observing Eq. \eqref{Eq:UAP}, where the function $G(\mathbf{x})$ produces scalar outputs in the real number field $\mathbb{R}$, the scenario naturally extends when $G(\mathbf{x})$ maps to $\mathbb{R}^m$, requiring approximation in each dimension. To accommodate such multi-dimensional outputs, we simply need to extend Eq. \eqref{Eq:UAP}: the transformation matrix $\mathbf{W}_j$ is revised to the space $\mathbb{R}^{n \times m}$, the bias term $\theta_j$ is recast as a vector in $\mathbb{R}^m$, and $\alpha_j$ is reshaped into a matrix. We call the parameters $\mathbf{W}_j$ and $\alpha_j$ in UAT as weight and $\theta_j$ as bias.

\subsection{The Problems in Designing parallel Network Based on UAT}
\label{section:The Problem in Designing parallel Network Based on UAT}

\begin{figure}[hbpt!]
\centering
\includegraphics[width=0.48\textwidth]{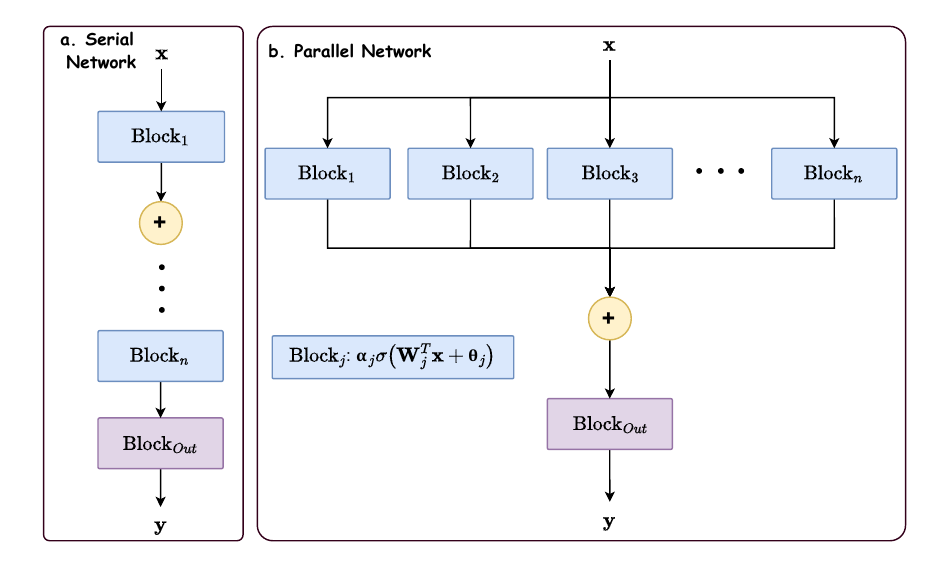}
\caption{The general description of the serial network and parallel network based on UAT.}
\label{fig:para-net}
\end{figure}

From Eq. \eqref{Eq:UAP}, we can observe a significant characteristic: if we define different indices $j$ as the indices of layers and the corresponding layer's mathematical form is $\alpha_j \sigma\left(\mathbf{W}_j^{\mathrm{T}} \mathbf{x}+\theta_j\right)$, it is evident that the inputs, outputs and parameters between different layers do not interact with each other. Therefore, based on Eq. \eqref{Eq:UAP}, we can design networks where parameters across multiple layers are independent, achieving parallel computation, like Figure \ref{fig:para-net}. Figure \ref{fig:para-net}.a represents a typical sequential network, which clearly requires computation to proceed in order from front to back. In contrast, Figure \ref{fig:para-net}.b depicts a parallel network, where there is no parameter interaction between different blocks, allowing them to be considered independent and thus suitable for parallel computation. The blocks in this network should satisfy the formula of $\alpha_j \sigma\left(\mathbf{W}_j^{\mathrm{T}} \mathbf{x}+\theta_j\right)$. Clearly, under this condition, the sum of the multiple blocks' outputs adheres to the UAT theory.

If we design the parallel network solely based on Eq. \eqref{Eq:UAP}, the network will face a limitation: Bad performance. This network format is a single layer perception network \cite{MinskySeymour1969PerceptronsAI} and it is out of date because of poor performance. The fundamental reason is that once training is complete: the function that can be approximated by the network is fixed, as the parameters in the corresponding UAT are fixed. Although UAT has powerful approximation ability, the real world is complex. It is impossible to approximate the world using a fix function.

\section{DUAT to Parallel Network}
\label{section:UAT for Parallel Network}
In Section \ref{section:The Problem in Designing parallel Network Based on UAT}, we have given the drawback of designing parallel networks based on UAT straightforwardly. So to ensure the network can dynamically approximate the world, we adopt the DUAT. DUAT's general mathematical form is the same as UAT, but some or all parameters in DUAT \cite{wang2024universalapproximationtheorybasic1, wang2024universalapproximationtheorybasic} can dynamically change based on input. This allows the network to dynamically adapt and fit the function based on the input.

\subsection{Introduction of DUAT}
\label{section:Introduction of DUAT}
The DUAT was introduced in \citet{wang2024universalapproximationtheorybasic1, wang2024universalapproximationtheorybasic}, which can be seen as an advanced form of UAT based on the residual structure. To introduce DUAT, we first explain the Matrix-Vector Method \cite{wang2024universalapproximationtheorybasic1, wang2024universalapproximationtheorybasic}. In neural networks, there are a variety of fundamental operations, such as convolution, pooling, multi-head attention (MHA), and feed-forward networks (FFN), they are represented  primarily by engineer format rather than expressed in a unified mathematical form.

\begin{figure}[htbp!]
\centering
\includegraphics[width=0.25\textwidth]{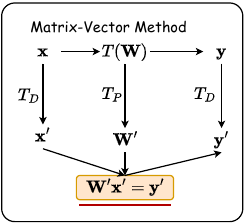}
\caption{The process of Transformer module.}
\label{fig:MVM}
\end{figure}

To address this, \citet{wang2024universalapproximationtheorybasic1, wang2024universalapproximationtheorybasic} introduced the Matrix-Vector Method, a technique that standardizes these operations within a unified mathematical framework, as shown in Figure \ref{fig:MVM}. In this context, $\mathbf{x}$ and $\mathbf{y}$ denote the input and output of a basic operation, with $ T(\mathbf{W}) $ representing various transformations, such as convolution, and $\mathbf{W}$ corresponding to the parameters of the transformation $ T $. The transformations $ T_{D} $ and $ T_{P} $ are applied to convert $\mathbf{x}$, $\mathbf{y}$, and $\mathbf{W}$ into column vectors $\mathbf{x}'$ and $\mathbf{y}'$, and a parameter matrix $\mathbf{W}'$, ensuring that the equation $\mathbf{W}'\mathbf{x}' = \mathbf{y}'$ is satisfied.

To clearly distinguish these transformed variables, we mark them with a prime symbol, for instance, $\mathbf{x} \mapsto \mathbf{x}'$. With this standardization, we can express the operations of multilayer networks as a cohesive mathematical form, allowing for a unified representation across different network structures.

\begin{figure}[htbp!]
\centering
\includegraphics[width=0.45\textwidth]{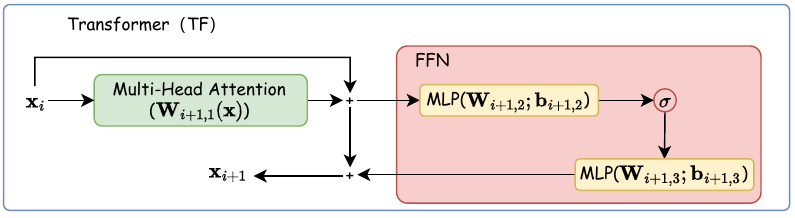}
\caption{The process of Transformer module.}
\label{fig:TF}
\end{figure}

Following this approach, Figure \ref{fig:TF} provides an example of the typical operations within a module in a Transformer model. According to the Matrix-Vector Method, MHA can be represented as \cite{wang2024universalapproximationtheorybasic1}:
\begin{equation}
MHA(\mathbf{x}_{i}) \mapsto \mathbf{W}'_{i+1,1}\mathbf{x}'_{i}
\label{eq:MHA}
\end{equation}

FFN can be expressed as \cite{wang2024universalapproximationtheorybasic1}:
\begin{equation}
FFN(\mathbf{x}_{i}) \mapsto  \mathbf{W}'_{i+1,3}\sigma (\mathbf{W}'_{i+1,2}\mathbf{x}'_{i}+\mathbf{b}'_{i+1,2})+\mathbf{b}'_{i+1,3}\\
\label{eq:FFN}
\end{equation}

According to Eq. \eqref{eq:MHA} and Eq. \eqref{eq:FFN}, the output $\mathbf{x}'_{1}$ of a single-layer Transformer can be expressed as Eq. \eqref{eq:TF-UAT-1-layer} (see \textcolor{blue}{Appendix A} for more details). For convenience, the mathematical form of Eq. \eqref{eq:TF-UAT-1-layer} can be considered equivalent to the two terms $\alpha_j \sigma\left(\mathbf{W}_j^{\mathrm{T}} \mathbf{x}+\theta_j\right)$ in Eq. \eqref{Eq:UAP}. 

\begin{equation}
\centering
\begin{aligned}
\mathbf{x}'_{1} &= (\mathbf{W}'_{1,1}\mathbf{x}'_{0} + \mathbf{b}'_{i,3})+ \mathbf{W}'_{1,3}\sigma (\mathbf{W}'_{1,2}\mathbf{x}'_{1} + \mathbf{b}'_{1,2}) 
\end{aligned}
\label{eq:TF-UAT-1-layer}
\end{equation}

Using Figure \ref{fig:TF} as a reference for a typical module in a multi-layer Transformer network, we can derive that a multi-layer Transformer is effectively a DUAT function, which can be expressed as:

\begin{equation}
\begin{aligned}
\mathbf{x}_{i+1}=(\mathbf{W}_{i+1,1}'\mathbf{x}_0+\mathbf{b}_{i+1,1}')+\sum_{j=1}^{i+1}\mathbf{W}'_{j,3}\sigma (\mathbf{W}'_{j,2}\mathbf{x}'_{0}+\mathbf{b}'_{j,2})
\end{aligned}
\label{eq:TF}
\end{equation}

For the case where $ i = 0 $, the parameters are defined as follows: $ \mathbf{W}'_{1,1} = \mathbf{W}'_{1,1} $, $ \mathbf{b}'_{1,3} = \mathbf{b}'_{1,3} $, $ \mathbf{W}'_{1,3} = \mathbf{W}'_{1,3} $, $ \mathbf{W}'_{1,2} = \mathbf{W}'_{1,2} \mathbf{W}'_{1,1} $, and $ \mathbf{b}'_{1,2} = \mathbf{b}'_{1,2} $. For cases where $ i \geq 1 $, we define the parameter updates as follows. For each $ j = 1, 2, \ldots, i $, the updates are: $ \mathbf{W}'_{j+1,1} = \mathbf{W}'_{j+1,1} \mathbf{W}_{j,1} $, $ \mathbf{b}'_{j+1,3} = \mathbf{W}'_{j+1,1} \mathbf{b}_{j,3} + \mathbf{b}'_{j+1,3} $, $ \mathbf{W}'_{j+1,2} = \mathbf{W}'_{j+1,2} \mathbf{W}_{j,1}$, and $ \mathbf{W}'_{j+1,3} = \mathbf{W}'_{j+1,1} \mathbf{W}'_{j,3} $. Additionally, for $ j = 2, \ldots, i + 1 $, the bias terms $ \mathbf{b}'_{j,2} $ are updated according to:

\begin{equation}
\begin{aligned}
\mathbf{b}'_{j,2} =& (\mathbf{W}'_{j,2} \mathbf{b}'_{j-1,3} + \mathbf{b}'_{j,2}) \\
+& \mathbf{W}'_{j,2} \sum_{k=1}^{j-1} \mathbf{W}'_{k,3} \sigma (\mathbf{W}'_{k,2} \mathbf{x}'_{0} + \mathbf{b}'_{k,2}).
\end{aligned}
\label{eq:b_j2}
\end{equation}

Besides, parameters within the MHA mechanism also vary dynamically with the input. Therefore, in the DUAT associated with Transformers, the parameters $\mathbf{W}'_{j,1}$ and $\mathbf{W}'_{j,2}$ for $i+1 \geq j \geq 1$, and $\mathbf{W}'_{j,3}$ for $i+1 > j > 1$ in layer $i$, all adapt dynamically based on the input.

\subsection{Para-former}
\label{section:Para-former}

\begin{figure}[htbp!]
\centering
\includegraphics[width=0.35\textwidth]{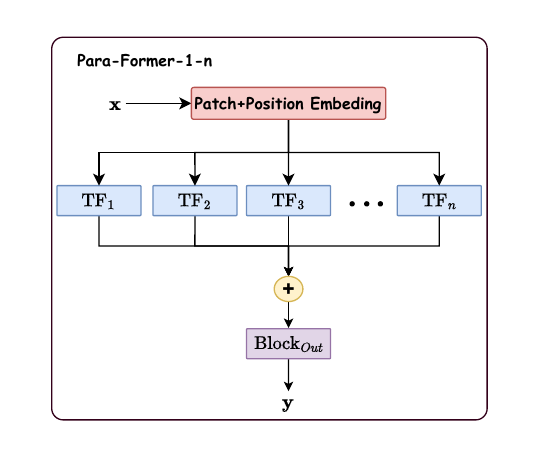}
\caption{The structure of the Para-Former-1-n represents Para-Former with 1 depth of n layers.}
\label{fig:para-former-1}
\end{figure}

\begin{figure}[htbp!]
\centering
\includegraphics[width=0.35\textwidth]{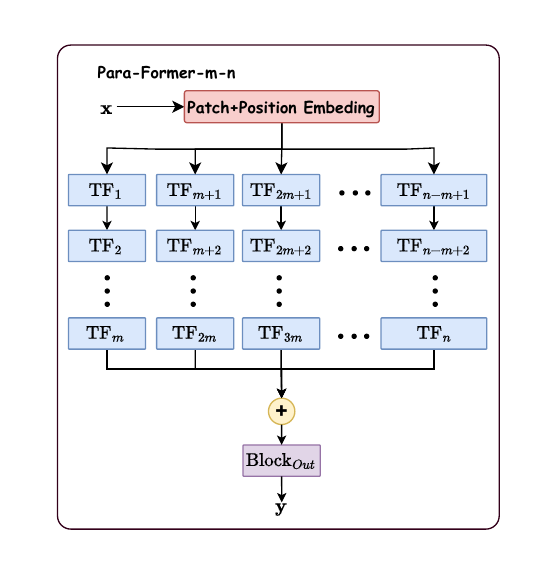}
\caption{The structure of the Para-Former-m-n represents Para-Former with m depth of n layers.}
\label{fig:para-former-2}
\end{figure}

In Section \ref{section:Introduction of DUAT}, we introduced DUAT using the Transformer model as an example. Following this rationale, we designed both single-depth and multi-depth parallel Transformer structures, as shown in Figures \ref{fig:para-former-1} and \ref{fig:para-former-2}. We selected the Transformer for this design due to its robust performance and established utility across both CV and NLP domains.

It is evident that Transformer-based models, whether single-layer or multi-layer, possess the capability to dynamically approximate functions based on input. According to Eq. \eqref{eq:TF}, the parameters responsible for this dynamic approximation in a single layer are $\mathbf{W}'_{1,1}$ and $\mathbf{W}'_{1,2}$. In multi-layer structures, both weights and bias terms exhibit dynamic approximation capabilities. Notably, the bias terms also leverage DUAT for approximation. Consequently, the input-output relationship and parameters in a multi-layer Transformer structure are inherently interrelated.

The advantage of the single-depth model lies in its reduced computation time; however, only the weight terms are dynamically changed. In contrast, the multi-depth model benefits from dynamic approximation in both its weights and multiple bias terms, with bias terms also employing DUAT \cite{wang2024universalapproximationtheorybasic1}. These approaches in networks design raises two key questions: is this network design feasible, and if so, which network performs better when the number of basic modules is identical? We will address these questions in Section \ref{section:Model Depth Effect}.

\subsection{Speed-up Ratio}
\label{section:Speed-up ratio}
We have proposed a network parallelization scheme, and in this section, we will compare the running speed of this parallelized network with that of traditional serial networks. It is important to note that due to differences in hardware and implementation methods, providing an objective comparison through experiments can be challenging. Therefore, we will focus on a theoretical analysis. Based on the network structure shown in Figure \ref{fig:para-former-1} and \ref{fig:para-former-2}, the inference time of Para-Former is only affected by its network depth. 

Assume we use a Para-Former with a depth of $M$ and a layer count of $L$, compared to a traditional Transformer network with a depth of $N$ layers, where $L >> N$. In this case, the running speed of the parallel network would be $N/M$ times faster than that of the serial network. This is because the speed of the parallel network is solely related to the depth of the Para-Former. As the serial network depth increases, the acceleration effect becomes more pronounced.

\section{Experiments}
\label{section:Experiments}
Our experimental approach is to first verify the feasibility of the parallel network theory and its adherence to the DUAT properties (The larger value of $N$, the more powerful the model becomes, resulting in better classification performance.) using several commonly used public image classification datasets (Section \ref{section:The Feasibility of Para-Former}). We then explore the impact of network depth on the Para-Former performance (Section \ref{section:Model Depth Effect}), followed by an investigation into the influence of the dataset (Section \ref{section:Dataset Effect}). We conducted experiments with various Para-Former models, where the naming convention uses the first number to indicate the depth and the second number to indicate the number of layers. For instance, Para-former-1-12 denotes a model with 1 depth and 12 layers.

\begin{table*}[htbp!]
    \centering
\begin{tabular}{cccccc} 
\toprule
\multicolumn{1}{l}{} & Cifar-10 & Cifar-100 & STL-10 & \multicolumn{1}{l}{OxfordIIITPet} & Flower-102  \\ 
\hline
Classes              & 10       & 100       & 10     & 37                                & 102         \\
Train set            & 47500    & 47500     & 4750   & 3,515                             & 969         \\
Val set              & 2500     & 47500     & 250    & 185                               & 51          \\
Test set             & 10000    & 10000     & 800    & 3669                              & 6149        \\
Per class            & 4750     & 4750      & 475    & 95                                & 9.5         \\
\bottomrule
\end{tabular}
\caption{This table presents the sizes of the training, validation, and test sets for the datasets CIFAR-10, CIFAR-100, STL-10, OxfordIIITPet, and Flower-102. It also provides the number of categories (Classes) and the average number of images per category for each dataset (Per class).}
\label{tab:data-info}
\end{table*}

\subsection{Dataset}
\label{section:Dataset}
In pursuit of a comprehensive understanding of the efficacy and practicality of parallel models, we meticulously selected several widely-acknowledged and representative image classification datasets for an integrated comparative analysis. These datasets encompass a diverse array of visual content and span various levels of difficulty, thereby furnishing us with a fertile ground to probe the performance boundaries of parallel models. Specifically, our study centers on the following pivotal datasets: CIFAR-10, CIFAR-100 \cite{Krizhevsky2009LearningML}, STL-10 \cite{Coates2011AnAO}, Flowers-102 \cite{Nilsback2008AutomatedFC} and Oxford-IIIT Pets \cite{Parkhi2012CatsAD}. In Table \ref{tab:data-info}, we provide more details for these datasets.

\begin{table*}[htbp!]
\centering
\begin{tabular}{ccccc} 
\toprule
              & Para-F-1-1 & Para-F-1-6 & Para-F-1-12    & \multicolumn{1}{l}{Para-F-1-24}  \\ 
\hline
Cifar-10      & 55.95      & 64.90      & 66.88          & \textbf{71.53}                   \\
Cifar-100     & 30.67      & 38.84      & 41.41          & \textbf{46.96}                   \\
STL-10        & 44.41      & 51.57      & 55.96          & \textbf{59.53}                   \\
OxfordIIITPet & 19.62      & 20.95      & 22.07          & \textbf{22.73}                   \\
Flower-102    & 27.33      & 32.65      & \textbf{37.42} & 35.84                            \\
\bottomrule
\end{tabular}
\caption{This table shows the predictions of Para-former-1-1, Para-former-1-6, Para-former-1-12, Para-former-1-24 on the datasets CIFAR-10, CIFAR-100, STL-10, OxfordIIITPet, and Flower-102. We simply represent them as Para-F-1-1, Para-F-1-6, Para-F-1-12 and Para-F-1-24.}
\label{tab:1-depth-result}
\end{table*}

\subsection{The Feasibility of Para-Former}
\label{section:The Feasibility of Para-Former}
To validate the feasibility and characteristics of Para-Former, we conducted training and validation on several commonly used image classification datasets, with all models trained for 500 epochs. The experimental design involved preprocessing inputs using the common ViT techniques of patch extraction and positional encoding. We then used Transformer as the basic model architecture, setting all network depths to 1 and configuring Para-Former from single-layer to multi-layer setups, as shown in Figures \ref{fig:para-former-1} and \ref{fig:para-former-2} (We evaluated the models' performance using the accuracy metric, calculated as $ \text{accuracy} = 100 \times\frac{ \text{correct}}{\text{total}}$). A single-layer setup corresponds to a ViT with a depth of 1. The results of the models are presented in Table \ref{tab:1-depth-result}.

It is evident from Table \ref{tab:1-depth-result} that as the number of parallel network layers increases, the model accuracy improves progressively (Except for the result of Para-F-1-24 being worse than Para-F-1-12 on the Flower-102 dataset, this variation is within a reasonable range.). According to DUAT theory, increasing the network layers is equivalent to increasing $N$ in Eq. \eqref{Eq:UAP}, thereby enhancing the approximation capability of DUAT. The results in Table \ref{tab:1-depth-result} are in complete agreement with DUAT. Therefore, we believe that this parallel network approach is entirely feasible.

\begin{table*}[htbp!]
\centering
\begin{tabular}{ccccc} 
\toprule
              & Para-F-6-6     & Para-F-2-24 & Para-F-3-24 & \multicolumn{1}{l}{Para-F-6-24}  \\ 
\hline
Cifar-10      & 79.39          & 74.75       & 77.21       & \textbf{79.92}                   \\
Cifar-100     & 55.18          & 51.04       & 52.65       & \textbf{56.15}                   \\
STL-10        & 62.81          & 61.60       & 62.75       & \textbf{64.30}                   \\
OxfordIIITPet & \textbf{26.05} & 23.65       & 26.02       & 25.92                            \\
Flower-102    & 37.97          & ~39.17      & 40.12       & \textbf{40.51}                   \\
\bottomrule
\end{tabular}
\caption{This table shows the predictions of Para-former-6-6, Para-former-2-24, Para-former-3-24, Para-former-6-24 on the datasets CIFAR-10, CIFAR-100, STL-10, OxfordIIITPet, and Flower-102. }
\label{tab:different-depth}
\end{table*}

However, the overall prediction accuracy in Table \ref{tab:1-depth-result} is quite low. What could be the reasons for this? We believe there are two main factors affecting the model's predictive performance:

1. Model Fitting Capability: This is influenced by the model’s size and design, particularly the value of $N$ in the DUAT and the degrees of freedom of the parameters. A larger $N$ enhances the DUAT's ability to fit functions. The degrees of freedom refer to the number of parameters in the DUAT that are influenced by the input. For instance, the degrees of freedom in Eq. \eqref{eq:TF-UAT-1-layer} is 2 (see \textcolor{blue}{Appendix B}). The degrees of freedom represent the model's ability to dynamically fit corresponding functions based on the input. Simply increasing $N$ can improve the model's ability to fit specific functions, but images are usually diverse. Thus, the network must have the capability to dynamically fit different functions based on the input. Higher degrees of freedom allow the model to fit multiple different functions according to the input. Additionally, the degrees of freedom vary within the Transformer: they mainly consist of weights and biases. The degrees of freedom for weights stem from the MHA, which can be simply viewed as a generated parameters based the covariance matrix of the input. The degrees of freedom for biases come from residual connections, which can dynamically fit the bias terms using DUAT. The deeper the residual layers, the more DUAT layers correspond to the bias terms, leading to stronger dynamic fitting capabilities. As a result, a shallow serial network can achieve the fitting ability of a multi-layer parallel network with a depth of 1. We will explore this issue further in Section \ref{section:Model Depth Effect}.

2. Data Influence: Data is another crucial factor in deep learning training. Since natural images are typically diverse, limited data can lead to overfitting. For example, in a cat-dog classification task, if dogs in the training set mostly appear against backgrounds of blue skies and grass, these backgrounds might be mistakenly considered features of dogs. If cats rarely appear in such backgrounds in the training set but do in the test set, they might be misclassified as dogs. Training on a large, diverse dataset means common backgrounds frequently appear with various objects or animals, such as cats, dogs, and elephants alongside blue skies, white clouds, houses, and trees. Therefore the model will learn background information unconsciously when doing classification tasks. Then fine-tuning on a smaller cat-dog dataset will yield better results because the background won't interfere with the recognition of dog features. Without sufficient data, even a network with strong fitting capabilities cannot perform well and is prone to overfitting. We will consider this issue in Section \ref{section:Dataset Effect}.

\subsection{Model Depth Effect}
\label{section:Model Depth Effect}

Based on the conclusions from Section \ref{section:The Feasibility of Para-Former}, this section explores the impact of network depth on prediction performance. According to equations \eqref{eq:TF-UAT-1-layer} and \eqref{eq:TF}, the depth of the model primarily affects the degrees of freedom of the model parameters within the DUAT. Therefore, we designed different versions of the Para-Former model with 24 layers and depths of 2, 3, and 6, comparing their predictive performance against a ViT with a depth of 6. The results are shown in Table \ref{tab:different-depth}. The degrees of freedom for Para-F-1-24, Para-F-6-6, Para-F-2-24, Para-F-3-24, and Para-F-6-24 are 48, 15, 72, 72, and 60, respectively (see \textcolor{blue}{Appendix B}).

Why does Para-F-6-6, despite having the lowest degrees of freedom, only perform worse than Para-F-6-24? This is because the degrees of freedom encompass both the freedom of weights and biases. The freedom of weights can be understood as linear transformations derived from inputs covariance, whereas the freedom of biases results from the approximation by multiple layers of DUAT, which allows for stronger dynamic fitting capabilities. The deeper the layers, the larger number of $N$ in DUAT for bias term, and the stronger the approximation ability. Thus, Para-F-6-24 performs the best, followed by Para-F-6-6, as the number of DUAT layer of their bias terms is largest (see \textbf{Appendix B}).

In summary, deeper models have stronger approximation capabilities, but there is a limit to how deep a model can go, necessitating a balance between network depth and layer count. 

\subsection{Dataset Effect}
\label{section:Dataset Effect}

\begin{table*}[hbpt!]
\centering
\begin{tabular}{lccclc} 
\toprule
                                 & Cifar-10 & Cifar-100 & STL-10 & OxfordIIITPet             & Flower-102  \\ 
\hline
\multicolumn{1}{c}{Para-F-12-12} & 98.62    & 91.36     & 99.47  & \multicolumn{1}{c}{91.87} & 96.89       \\
\bottomrule
\end{tabular}
\caption{This table shows the results of fine-tuning using Google's official pre-trained parameters of ViT. The pre-trained parameters were obtained from training the ViT model on the ImageNet dataset.}
\label{tab:finetune}
\end{table*}

After increasing the depth of the Para-Former, the results have improved to some extent. However, as shown in Table \ref{tab:different-depth}, almost all models encounter a performance bottleneck, especially on the OxfordIIITPet and Flower-102 datasets. What is the fundamental reason for this poor performance? This is likely because these datasets are relatively small and exhibit high data diversity. In this section, we will solve this problem.

First, the experimental results are based on validated ViT models (Para-F-1-1 and Para-Former-6-6 are equal to ViT with 1 and 6 depth.), whose performance is similar to that of the Para-Former. Given that ViT models are widely validated and are among the commonly used model architectures, their effectiveness is beyond doubt. So why is the performance so poor? The fundamental reason lies in the data itself: the model's performance is typically directly related to the amount of available data. Take the Flower-102 dataset as an example, which contains 102 classes of flowers but only 1,020 training images, with some reserved for validation. This is far from sufficient for training deep learning models. Why is the performance even worse on the OxfordIIITPet dataset, which has a higher average number of images per class? This is due to the diversity of the data (i.e., the similarity of features within the same class) and the variability of the background.

\begin{figure*}[htbp!]
\centering
\includegraphics[width=0.9\textwidth]{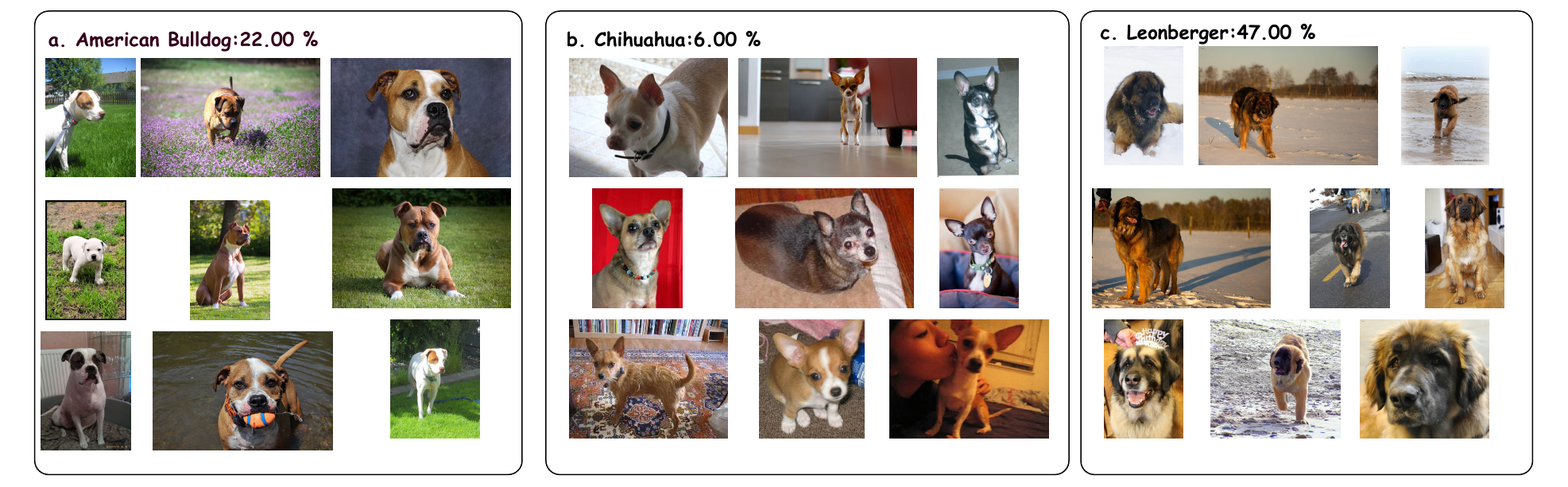}
\caption{OxfordIIITPet dataset's accuracy on different classes.}
\label{fig:pet}
\end{figure*}

\begin{figure*}[htbp!]
\centering
\includegraphics[width=0.9\textwidth]{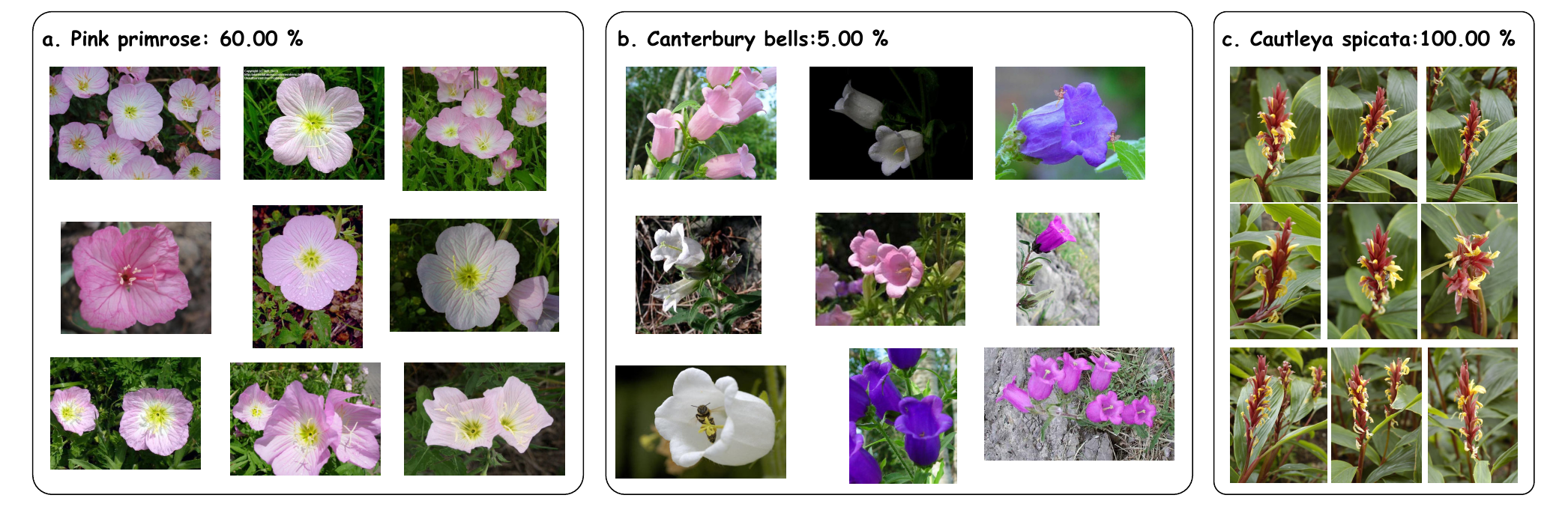}
\caption{Flower-102 dataset's accuracy on different classes.}
\label{fig:Flower102}
\end{figure*}

Figures \ref{fig:pet} and \ref{fig:Flower102} show some results from the OxfordIIITPet and Flower-102 datasets. Both figures indicate that for categories with diverse backgrounds and inherent feature diversity, the prediction results are poor, as shown in Figures \ref{fig:pet}.b and \ref{fig:Flower102}.b. In contrast, when the background is consistent but the object morphology is diverse, the results are relatively better, as shown in Figures \ref{fig:pet}.a and \ref{fig:Flower102}.a. The best results are achieved when both the background and object morphology are minimal, as shown in Figures \ref{fig:pet}.c and \ref{fig:Flower102}.c. Although the Flower-102 dataset has fewer images, its backgrounds are more consistent, and the differences between species are smaller. Conversely, animals in the OxfordIIITPet dataset can appear in various forms, whereas flowers in the Flower-102 dataset are usually photographed from relatively fixed angles, leading to better performance.

\begin{figure*}[h]
\centering
\includegraphics[width=0.8\textwidth]{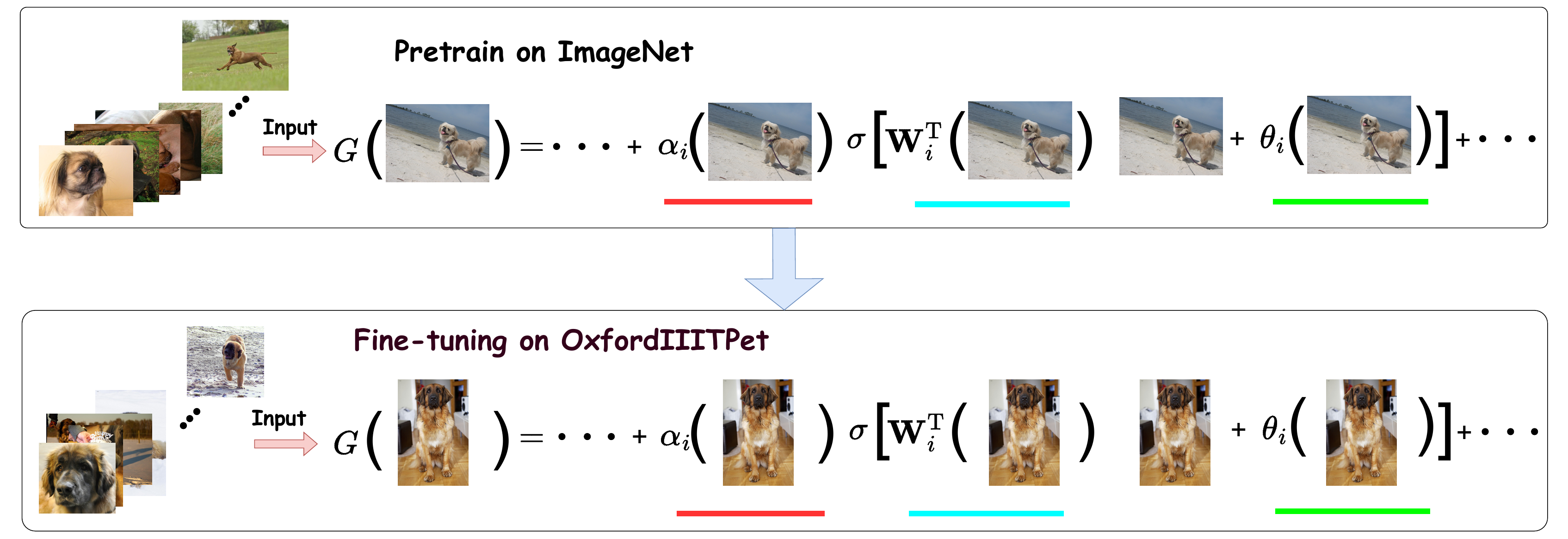}
\caption{The fine-tuning process from the perspective of UAT.}
\label{fig:fine-tune}
\vspace{-1em}
\end{figure*}

Does this mean that deep learning models are not powerful enough or lack generalization and reasoning capabilities? Quite the opposite. With such limited data, deep learning models achieved about 25\% accuracy on OxfordIIITPet and about 40\% on Flower-102, actually demonstrating the strength and generalization ability of deep learning models. A common misconception about deep learning is that even if people are unfamiliar with certain species, by careful examination, they can accurately distinguish between OxfordIIITPet and Flower-102 images, leading to the belief that deep learning lacks generalization or human-like reasoning capabilities. This view is flawed. Suppose the human brain operates like a DUAT model; it can automatically ignore background information before seeing these images because such backgrounds are frequently encountered in daily life, and the DUAT in the brain has already fitted this information. Therefore, people only need to focus on specific information and then use this information to train the brain to identify similarities and differences. For example, a 10-year-old child's brain, observing the world at 10 frames per second \cite{read2000restoration}, would have processed over $1.5 \times 10^{9}$ high-resolution images by age 10 \cite{lu2017high}, many containing common objects and backgrounds from daily life. Hence, it is unfair to criticize deep learning models for being unable to train on small datasets and for lacking universality and generalization ability.

Therefore, we should train on large datasets and then fine-tune on smaller ones. This is also the current method for deep learning training, demonstrating the powerful generalization capabilities of deep learning models, especially when there is some correlation between the datasets. Following this approach, we fine-tuned a pre-trained ViT model using only 10 epochs. The results, as shown in Table \ref{tab:finetune}, demonstrate significant improvements in prediction performance.

Figure \ref{fig:fine-tune} illustrates the rationale for fine-tuning. The goal, whether in training or prediction, is to obtain the target results based on the input. This can be understood as function fitting, expressed as $\mathbf{y}=\mathbf{G}(\mathbf{x}|\mathbf{A},\mathbf{W},\mathbf{\Theta})$, where $\mathbf{A}={\mathbf{\alpha}_1...\mathbf{\alpha}_N}$, $\mathbf{W}={\mathbf{W}_1...\mathbf{W}_N}$, and $\mathbf{\Theta}={\mathbf{\theta}_1...\mathbf{\theta}_N}$. The purpose of training is to determine $\mathbf{A}, \mathbf{W}, \mathbf{\Theta}$, which can dynamically change based on the input.

This means that if the background is similar but the objects are different, the resulting parameters will differ. If the model can make correct predictions, it implies that the model can distinguish between the background and the animals. If it could not be distinguished, it might incorrectly use the background as a feature of the animals, leading to the same classification. Moreover, if the model can correctly classify the animals, it indicates that the model can identify the features of the animals.

When the backgrounds are different, because natural backgrounds are generally similar, extensive training can differentiate background information. Thus, if the model can make right predictions, it suggests that the model has comprehensively grasped the animal features, even with significant environmental variations. Training on ImageNet follows this process. A new dataset is just a new series of combinations of various similar backgrounds and animals which have been seen in ImageNet. If the model has already learned sufficient background and animal information on ImageNet, it can dynamically fit the results based on the input. Therefore, only minor adjustments specific to new data are required during fine-tuning.

\section{Conclusion}
In this paper, we propose design guidelines for parallel neural networks based on DUAT, which, to our knowledge, is the first work to establish principles for parallel networks. Based on this theory, we designed a parallel network, Para-Former, and validated its effectiveness across several commonly used datasets. Through DUAT, we analyzed the characteristics of both serial and parallel networks. Serial networks with residual connections often have higher parameter flexibility, leading to better generalization capabilities. However, as network depth increases, inference time also grows. In contrast, parallel networks trade off computational resources for time efficiency, meaning their runtime is unaffected by the number of layers. This allows us to design multi-layer parallel networks that can replace serial networks, achieving faster inference. Additionally, we examined the impact of data on deep learning performance and explored the underlying reasons. From the DUAT perspective, training with large datasets essentially enhances a model’s capacity to fit diverse data.

\bibliography{iclr2025_conference}

\begin{thebibliography}{34}
\providecommand{\natexlab}[1]{#1}
\providecommand{\url}[1]{\texttt{#1}}
\expandafter\ifx\csname urlstyle\endcsname\relax
  \providecommand{\doi}[1]{doi: #1}\else
  \providecommand{\doi}{doi: \begingroup \urlstyle{rm}\Url}\fi

\bibitem[Achiam et~al.(2023)Achiam, Adler, Agarwal, Ahmad, Akkaya, Aleman,
  Almeida, Altenschmidt, Altman, Anadkat, et~al.]{achiam2023gpt}
Josh Achiam, Steven Adler, Sandhini Agarwal, Lama Ahmad, Ilge Akkaya,
  Florencia~Leoni Aleman, Diogo Almeida, Janko Altenschmidt, Sam Altman,
  Shyamal Anadkat, et~al.
\newblock Gpt-4 technical report.
\newblock \emph{arXiv preprint arXiv:2303.08774}, 2023.

\bibitem[Ben-Nun \& Hoefler(2018)Ben-Nun and
  Hoefler]{bennun2018demystifyingparalleldistributeddeep}
Tal Ben-Nun and Torsten Hoefler.
\newblock Demystifying parallel and distributed deep learning: An in-depth
  concurrency analysis, 2018.
\newblock URL \url{https://arxiv.org/abs/1802.09941}.

\bibitem[Brown et~al.(2020)Brown, Mann, Ryder, Subbiah, Kaplan, Dhariwal,
  Neelakantan, Shyam, Sastry, Askell, et~al.]{brown2020language}
Tom Brown, Benjamin Mann, Nick Ryder, Melanie Subbiah, Jared~D Kaplan, Prafulla
  Dhariwal, Arvind Neelakantan, Pranav Shyam, Girish Sastry, Amanda Askell,
  et~al.
\newblock Language models are few-shot learners.
\newblock \emph{Advances in neural information processing systems},
  33:\penalty0 1877--1901, 2020.

\bibitem[Coates et~al.(2011)Coates, Ng, and Lee]{Coates2011AnAO}
Adam Coates, A.~Ng, and Honglak Lee.
\newblock An analysis of single-layer networks in unsupervised feature
  learning.
\newblock In \emph{International Conference on Artificial Intelligence and
  Statistics}, 2011.
\newblock URL \url{https://api.semanticscholar.org/CorpusID:308212}.

\bibitem[Cybenko(1989)]{Cybenko1989ApproximationBS}
George~V. Cybenko.
\newblock Approximation by superpositions of a sigmoidal function.
\newblock \emph{Mathematics of Control, Signals and Systems}, 2:\penalty0
  303--314, 1989.
\newblock URL \url{https://api.semanticscholar.org/CorpusID:3958369}.

\bibitem[Dosovitskiy et~al.(2020)Dosovitskiy, Beyer, Kolesnikov, Weissenborn,
  Zhai, Unterthiner, Dehghani, Minderer, Heigold, Gelly, Uszkoreit, and
  Houlsby]{Dosovitskiy2020AnII}
Alexey Dosovitskiy, Lucas Beyer, Alexander Kolesnikov, Dirk Weissenborn,
  Xiaohua Zhai, Thomas Unterthiner, Mostafa Dehghani, Matthias Minderer, Georg
  Heigold, Sylvain Gelly, Jakob Uszkoreit, and Neil Houlsby.
\newblock An image is worth 16x16 words: Transformers for image recognition at
  scale.
\newblock \emph{ArXiv}, abs/2010.11929, 2020.
\newblock URL \url{https://api.semanticscholar.org/CorpusID:225039882}.

\bibitem[Elbtity et~al.(2024)Elbtity, Chandarana, and
  Zand]{Elbtity2024FlexTPUAF}
Mohammed~E. Elbtity, Peyton~S. Chandarana, and Ramtin Zand.
\newblock Flex-tpu: A flexible tpu with runtime reconfigurable dataflow
  architecture.
\newblock 2024.
\newblock URL \url{https://api.semanticscholar.org/CorpusID:271097592}.

\bibitem[Gale et~al.(2020)Gale, Zaharia, Young, and Elsen]{Gale2020SparseGK}
Trevor Gale, Matei~A. Zaharia, Cliff Young, and Erich Elsen.
\newblock Sparse gpu kernels for deep learning.
\newblock \emph{SC20: International Conference for High Performance Computing,
  Networking, Storage and Analysis}, pp.\  1--14, 2020.
\newblock URL \url{https://api.semanticscholar.org/CorpusID:219955899}.

\bibitem[Gou et~al.(2020)Gou, Yu, Maybank, and Tao]{Gou2020KnowledgeDA}
Jianping Gou, B.~Yu, Stephen~J. Maybank, and Dacheng Tao.
\newblock Knowledge distillation: A survey.
\newblock \emph{International Journal of Computer Vision}, 129:\penalty0 1789
  -- 1819, 2020.
\newblock URL \url{https://api.semanticscholar.org/CorpusID:219559263}.

\bibitem[Gu et~al.(2021)Gu, Lin, Kuo, and Cui]{Gu2021OpenvocabularyOD}
Xiuye Gu, Tsung-Yi Lin, Weicheng Kuo, and Yin Cui.
\newblock Open-vocabulary object detection via vision and language knowledge
  distillation.
\newblock In \emph{International Conference on Learning Representations}, 2021.
\newblock URL \url{https://api.semanticscholar.org/CorpusID:238744187}.

\bibitem[Guo et~al.(2021)Guo, Wang, and Cui]{Guo2021IntegerOnlyNN}
Qingyu Guo, Yuan Wang, and Xiaoxin Cui.
\newblock Integer-only neural network quantization scheme based on
  shift-batch-normalization.
\newblock \emph{ArXiv}, abs/2106.00127, 2021.
\newblock URL \url{https://api.semanticscholar.org/CorpusID:235266137}.

\bibitem[He et~al.(2015)He, Zhang, Ren, and Sun]{He2015DeepRL}
Kaiming He, X.~Zhang, Shaoqing Ren, and Jian Sun.
\newblock Deep residual learning for image recognition.
\newblock \emph{2016 IEEE Conference on Computer Vision and Pattern Recognition
  (CVPR)}, pp.\  770--778, 2015.
\newblock URL \url{https://api.semanticscholar.org/CorpusID:206594692}.

\bibitem[Hornik et~al.(1989)Hornik, Stinchcombe, and
  White]{Hornik1989MultilayerFN}
Kurt Hornik, Maxwell~B. Stinchcombe, and Halbert~L. White.
\newblock Multilayer feedforward networks are universal approximators.
\newblock \emph{Neural Networks}, 2:\penalty0 359--366, 1989.
\newblock URL \url{https://api.semanticscholar.org/CorpusID:2757547}.

\bibitem[Jacob et~al.(2017)Jacob, Kligys, Chen, Zhu, Tang, Howard, Adam, and
  Kalenichenko]{Jacob2017QuantizationAT}
Benoit Jacob, Skirmantas Kligys, Bo~Chen, Menglong Zhu, Matthew Tang, Andrew~G.
  Howard, Hartwig Adam, and Dmitry Kalenichenko.
\newblock Quantization and training of neural networks for efficient
  integer-arithmetic-only inference.
\newblock \emph{2018 IEEE/CVF Conference on Computer Vision and Pattern
  Recognition}, pp.\  2704--2713, 2017.
\newblock URL \url{https://api.semanticscholar.org/CorpusID:39867659}.

\bibitem[Jouppi et~al.(2017)Jouppi, Young, Patil, Patterson, Agrawal, Bajwa,
  Bates, Bhatia, Boden, Borchers, Boyle, luc Cantin, Chao, Clark, Coriell,
  Daley, Dau, Dean, Gelb, Ghaemmaghami, Gottipati, Gulland, Hagmann, Ho,
  Hogberg, Hu, Hundt, Hurt, Ibarz, Jaffey, Jaworski, Kaplan, Khaitan,
  Killebrew, Koch, Kumar, Lacy, Laudon, Law, Le, Leary, Liu, Lucke, Lundin,
  MacKean, Maggiore, Mahony, Miller, Nagarajan, Narayanaswami, Ni, Nix, Norrie,
  Omernick, Penukonda, Phelps, Ross, Ross, Salek, Samadiani, Severn, Sizikov,
  Snelham, Souter, Steinberg, Swing, Tan, Thorson, Tian, Toma, Tuttle,
  Vasudevan, Walter, Wang, Wilcox, and Yoon]{Jouppi2017IndatacenterPA}
Norman~P. Jouppi, Cliff Young, Nishant Patil, David Patterson, Gaurav Agrawal,
  Raminder Bajwa, Sarah Bates, Suresh Bhatia, Nan Boden, Al~Borchers, Rick
  Boyle, Pierre luc Cantin, Clifford Chao, Chris Clark, Jeremy Coriell, Mike
  Daley, Matt Dau, Jeffrey Dean, Ben Gelb, Taraneh Ghaemmaghami, Rajendra
  Gottipati, William Gulland, Robert Hagmann, C.~Richard Ho, Doug Hogberg, John
  Hu, Robert Hundt, Dan Hurt, Julian Ibarz, Aaron Jaffey, Alek Jaworski,
  Alexander Kaplan, Harshit Khaitan, Daniel Killebrew, Andy Koch, Naveen Kumar,
  Steve Lacy, James Laudon, James Law, Diemthu Le, Chris Leary, Zhuyuan Liu,
  Kyle Lucke, Alan Lundin, Gordon MacKean, Adriana Maggiore, Maire Mahony,
  Kieran Miller, Rahul Nagarajan, Ravi Narayanaswami, Ray Ni, Kathy Nix, Thomas
  Norrie, Mark Omernick, Narayana Penukonda, Andy Phelps, Jonathan Ross, Matt
  Ross, Amir Salek, Emad Samadiani, Chris Severn, Gregory Sizikov, Matthew
  Snelham, Jed Souter, Dan Steinberg, Andy Swing, Mercedes Tan, Gregory
  Thorson, Bo~Tian, Horia Toma, Erick Tuttle, Vijay Vasudevan, Richard Walter,
  Walter Wang, Eric Wilcox, and Doe~Hyun Yoon.
\newblock In-datacenter performance analysis of a tensor processing unit.
\newblock \emph{2017 ACM/IEEE 44th Annual International Symposium on Computer
  Architecture (ISCA)}, pp.\  1--12, 2017.
\newblock URL \url{https://api.semanticscholar.org/CorpusID:4202768}.

\bibitem[Krizhevsky(2009)]{Krizhevsky2009LearningML}
Alex Krizhevsky.
\newblock Learning multiple layers of features from tiny images.
\newblock 2009.
\newblock URL \url{https://api.semanticscholar.org/CorpusID:18268744}.

\bibitem[Krizhevsky et~al.(2012)Krizhevsky, Sutskever, and
  Hinton]{Krizhevsky2012ImageNetCW}
Alex Krizhevsky, Ilya Sutskever, and Geoffrey~E. Hinton.
\newblock Imagenet classification with deep convolutional neural networks.
\newblock \emph{Communications of the ACM}, 60:\penalty0 84 -- 90, 2012.
\newblock URL \url{https://api.semanticscholar.org/CorpusID:195908774}.

\bibitem[Lin et~al.(2023)Lin, Tang, Tang, Yang, Dang, and Han]{Lin2023AWQAW}
Ji~Lin, Jiaming Tang, Haotian Tang, Shang Yang, Xingyu Dang, and Song Han.
\newblock Awq: Activation-aware weight quantization for on-device llm
  compression and acceleration.
\newblock In \emph{Conference on Machine Learning and Systems}, 2023.
\newblock URL \url{https://api.semanticscholar.org/CorpusID:258999941}.

\bibitem[Lu et~al.(2017)Lu, Gu, Wang, and Zhang]{lu2017high}
Jing Lu, Boyu Gu, Xiaolin Wang, and Yuhua Zhang.
\newblock High-speed adaptive optics line scan confocal retinal imaging for
  human eye.
\newblock \emph{PloS one}, 12\penalty0 (3):\penalty0 e0169358, 2017.

\bibitem[Lu et~al.(2022)Lu, Zhong, and Yang]{Lu2022AHA}
Wei Lu, Ma~Zhong, and Chaojie Yang.
\newblock A hybrid asymmetric integer-only quantization method of neural
  networks for efficient inference.
\newblock \emph{2022 5th International Conference on Pattern Recognition and
  Artificial Intelligence (PRAI)}, pp.\  1326--1330, 2022.
\newblock URL \url{https://api.semanticscholar.org/CorpusID:252720487}.

\bibitem[Ma et~al.(2023)Ma, Fang, and Wang]{Ma2023LLMPrunerOT}
Xinyin Ma, Gongfan Fang, and Xinchao Wang.
\newblock Llm-pruner: On the structural pruning of large language models.
\newblock \emph{ArXiv}, abs/2305.11627, 2023.
\newblock URL \url{https://api.semanticscholar.org/CorpusID:258823276}.

\bibitem[MinskySeymour(1969)]{MinskySeymour1969PerceptronsAI}
Marvin MinskySeymour.
\newblock Perceptrons: An introduction to computational geometry.
\newblock 1969.
\newblock URL \url{https://api.semanticscholar.org/CorpusID:5400596}.

\bibitem[Nilsback \& Zisserman(2008)Nilsback and
  Zisserman]{Nilsback2008AutomatedFC}
Maria-Elena Nilsback and Andrew Zisserman.
\newblock Automated flower classification over a large number of classes.
\newblock \emph{2008 Sixth Indian Conference on Computer Vision, Graphics \&
  Image Processing}, pp.\  722--729, 2008.
\newblock URL \url{https://api.semanticscholar.org/CorpusID:15193013}.

\bibitem[Parkhi et~al.(2012)Parkhi, Vedaldi, Zisserman, and
  Jawahar]{Parkhi2012CatsAD}
Omkar~M. Parkhi, Andrea Vedaldi, Andrew Zisserman, and C.~V. Jawahar.
\newblock Cats and dogs.
\newblock \emph{2012 IEEE Conference on Computer Vision and Pattern
  Recognition}, pp.\  3498--3505, 2012.
\newblock URL \url{https://api.semanticscholar.org/CorpusID:383200}.

\bibitem[Radford \& Narasimhan(2018)Radford and
  Narasimhan]{Radford2018ImprovingLU}
Alec Radford and Karthik Narasimhan.
\newblock Improving language understanding by generative pre-training.
\newblock 2018.
\newblock URL \url{https://api.semanticscholar.org/CorpusID:49313245}.

\bibitem[Read \& Meyer(2000)Read and Meyer]{read2000restoration}
Paul Read and Mark-Paul Meyer.
\newblock \emph{Restoration of motion picture film}.
\newblock Elsevier, 2000.

\bibitem[Sergeev \& Balso(2018)Sergeev and Balso]{Sergeev2018HorovodFA}
Alexander Sergeev and Mike~Del Balso.
\newblock Horovod: fast and easy distributed deep learning in tensorflow.
\newblock \emph{ArXiv}, abs/1802.05799, 2018.
\newblock URL \url{https://api.semanticscholar.org/CorpusID:3398835}.

\bibitem[Shallue et~al.(2018)Shallue, Lee, Antognini, Sohl-Dickstein, Frostig,
  and Dahl]{Shallue2018MeasuringTE}
Christopher~J. Shallue, Jaehoon Lee, Joseph~M. Antognini, Jascha~Narain
  Sohl-Dickstein, Roy Frostig, and George~E. Dahl.
\newblock Measuring the effects of data parallelism on neural network training.
\newblock \emph{ArXiv}, abs/1811.03600, 2018.
\newblock URL \url{https://api.semanticscholar.org/CorpusID:53214190}.

\bibitem[Shi et~al.(2016)Shi, Cao, Zhang, Li, and Xu]{Shi2016EdgeCV}
Weisong Shi, Jie Cao, Quan Zhang, Youhuizi Li, and Lanyu Xu.
\newblock Edge computing: Vision and challenges.
\newblock \emph{IEEE Internet of Things Journal}, 3:\penalty0 637--646, 2016.
\newblock URL \url{https://api.semanticscholar.org/CorpusID:4237186}.

\bibitem[Simonyan \& Zisserman(2014)Simonyan and Zisserman]{Simonyan2014VeryDC}
Karen Simonyan and Andrew Zisserman.
\newblock Very deep convolutional networks for large-scale image recognition.
\newblock \emph{CoRR}, abs/1409.1556, 2014.
\newblock URL \url{https://api.semanticscholar.org/CorpusID:14124313}.

\bibitem[Touvron et~al.(2023)Touvron, Lavril, Izacard, Martinet, Lachaux,
  Lacroix, Rozi{\`e}re, Goyal, Hambro, Azhar, et~al.]{touvron2023llama}
Hugo Touvron, Thibaut Lavril, Gautier Izacard, Xavier Martinet, Marie-Anne
  Lachaux, Timoth{\'e}e Lacroix, Baptiste Rozi{\`e}re, Naman Goyal, Eric
  Hambro, Faisal Azhar, et~al.
\newblock Llama: Open and efficient foundation language models.
\newblock \emph{arXiv preprint arXiv:2302.13971}, 2023.

\bibitem[Wang \& Li(2024{\natexlab{a}})Wang and
  Li]{wang2024universalapproximationtheorybasic}
Wei Wang and Qing Li.
\newblock Universal approximation theory: The basic theory for deep
  learning-based computer vision models, 2024{\natexlab{a}}.
\newblock URL \url{https://arxiv.org/abs/2407.17480}.

\bibitem[Wang \& Li(2024{\natexlab{b}})Wang and
  Li]{wang2024universalapproximationtheorybasic1}
Wei Wang and Qing Li.
\newblock Universal approximation theory: The basic theory for large language
  models, 2024{\natexlab{b}}.
\newblock URL \url{https://arxiv.org/abs/2407.00958}.

\bibitem[Xia et~al.(2023)Xia, Gao, Zeng, and Chen]{Xia2023ShearedLA}
Mengzhou Xia, Tianyu Gao, Zhiyuan Zeng, and Danqi Chen.
\newblock Sheared llama: Accelerating language model pre-training via
  structured pruning.
\newblock \emph{ArXiv}, abs/2310.06694, 2023.
\newblock URL \url{https://api.semanticscholar.org/CorpusID:263830786}.

\end{thebibliography}
\bibliographystyle{iclr2025_conference}

\appendix
\clearpage
\appendix
\setcounter{figure}{0}
\setcounter{table}{0}

\section{The DUAT Format of single layer ViT}

In this section, we will demonstrate that the DUAT form of a single-layer layer ViT . Firstly, the basic form of a Transformer is shown in Figure 2 of main text. We represent the MHA and FFN in the Transformer in matrix-vector form as Eq. \eqref{eq:MHA-MV} and Eq. \eqref{eq:FFN-MV}, respectively (for specific details, refer to DUAT2LLMs). To distinguish between the operations and variables in the Transformer and those in the matrix-vector form, we add a prime ($'$) symbol to the operations and variables in the matrix-vector form. It is important to emphasize that the parameters of the MHA are influenced by the input. 


\begin{equation}
MHA'(\mathbf{x}'_{i}) \mapsto \mathbf{W}'_{i+1,1}\mathbf{x}'_{i}
\label{eq:MHA-MV}
\end{equation}

\begin{equation}
FFN'(\mathbf{x}'_{i}) \mapsto \mathbf{W}'_{i+1,3}\sigma (\mathbf{W}'_{i+1,2}\mathbf{x}'_{i}+\mathbf{b}'_{i+1,2})+\mathbf{b}'_{i+1,3}\\
\label{eq:FFN-MV}
\end{equation}

Next, the result of a single-layer Transformer $\mathbf{x}'_{1}$ is: 

\begin{equation}
\centering
\begin{aligned}
\mathbf{x}'_{1} & =\mathbf{W}'_{1,1}\mathbf{x}'_{0}+\mathbf{W}'_{1,3}\sigma [\mathbf{W}'_{1,2}(\mathbf{W}'_{1,1}\mathbf{x}'_{0})+\mathbf{b}'_{1,2}]+\mathbf{b}'_{1,3}\\
&=\mathbf{W}'_{1,1}\mathbf{x}'_{0}+\mathbf{W}'_{1,3}\sigma (\underline{ \mathbf{W}'_{1,2}\mathbf{W}'_{1,1}}\mathbf{x}'_{0}+\mathbf{b}'_{1,2})+\mathbf{b}'_{1,3}\\
\end{aligned}
\label{eq:TF-app-1}
\end{equation}

\section{The Degree of Freedom of Different Layers Transformer}
According to DUAT2LLMs, the DUAT format of a multi-layer ViT can be written as: 

\begin{equation}
\begin{aligned}
\mathbf{x}_{i+1}=(\mathbf{W}_{i+1,1}'\mathbf{x}_0+\mathbf{b}_{i+1,1})+\sum_{j=1}^{i+1}\mathbf{W}'_{j,3}\sigma (\mathbf{W}'_{j,2}\mathbf{x}'_{0}+\mathbf{b}'_{j,2})
\end{aligned}
\label{eq:TF-app}
\end{equation}

When $ i = 0 $, the parameters are defined as follows: $ \mathbf{W}'_{1,1} = \mathbf{W}'_{1,1} $, $ \mathbf{b}'_{1,3} = \mathbf{b}'_{1,3} $, $ \mathbf{W}'_{1,3} = \mathbf{W}'_{1,3} $, $ \mathbf{W}'_{1,2} = \mathbf{W}'_{1,2} \mathbf{W}'_{1,1} $, and $ \mathbf{b}'_{1,2} = \mathbf{b}'_{1,2} $. For cases where $ i \geq 1 $, we define the parameters iteratively. For each $ j = 1, 2, \ldots, i $, the updates are as follows: $ \mathbf{W}'_{j+1,1} = \mathbf{W}'_{j+1,1} \mathbf{W}_{j,1} $, $ \mathbf{b}'_{j+1,3} = \mathbf{W}'_{j+1,1} \mathbf{b}_{j,1} + \mathbf{b}'_{j+1,3} $, $ \mathbf{W}'_{j+1,2} = \mathbf{W}'_{j+1,2} \mathbf{W}_{j,1} $, and $ \mathbf{W}'_{j,3} = \mathbf{W}'_{i+1,1} \mathbf{W}'_{i,1} \cdots \mathbf{W}'_{j,3} $. Additionally, for $ j = 1, 2, \ldots, i + 1 $, the bias terms $ \mathbf{b}'_{j,2} $ are updated according to:
\begin{equation}
\begin{aligned}
\mathbf{b}'_{j,2} =& (\mathbf{W}'_{j,2} \mathbf{b}'_{j-1,3} + \mathbf{b}'_{j,2}) \\
+& \mathbf{W}'_{j,2} \sum_{k=1}^{j-1} \mathbf{W}'_{k,3} \sigma (\mathbf{W}'_{k,2} \mathbf{x}'_{0} + \mathbf{b}'_{k,2}).
\end{aligned}
\label{eq:b_j2-app}
\end{equation}

The term $\mathbf{b}'_{j,2}$ is approximated by a $j-1$-layer of DUAT with $\mathbf{x}_0$ as input. The parameters $\mathbf{W}'_{j,1}$ and $\mathbf{W}'_{j,2}$ for $i+1 \geq j \geq 1$, and $\mathbf{W}'_{j,3}$ for $i+1 > j > 1$ in layer $i$, all adapt dynamically based on the input. Because they are affected by the weight of MHA in each layer. 

Therefore, it is easy to deduce that the degrees of freedom for Transformers with 1, 2, 3, and 6 layers are 2, 6, 9, and 15, respectively. For Para-F-6-6, Para-F-2-24, Para-F-3-24, and Para-F-6-24, the degrees of freedom are shown in Table \ref{tab:DOF}. At first glance, it seems that the degrees of freedom for Para-F-1-24, Para-F-2-24, Para-F-3-24, and Para-F-6-24 all exceed that of Para-F-6-6. Why then does only Para-F-6-24 consistently outperform Para-F-6-6?

\begin{table*}[htbp!]
\centering
\begin{tabular}{lccclc} 
\toprule
                                      & Para-F-1-24 & Para-F-6-6 & Para-F-2-24 & Para-F-3-24             & Para-F-6-24  \\ 
\hline
\multicolumn{1}{c}{Degree of Freedom} & 2*24        & 15*1       & ~6*12        & \multicolumn{1}{c}{9*8} & 15*4         \\
\bottomrule
\end{tabular}
\caption{The degree of freedom of Para-F-1-24, Para-F-6-6, Para-F-2-24, Para-F-3-24, Para-F-6-24. We use the the degree of freedom of serial Transformer with the depth 1, 6, 2, and 3 to multiply the number of serial Transformer.}
\label{tab:DOF}
\end{table*}

The reason lies in the nature of the degrees of freedom. Taking $\mathbf{W}'_{2,2}$ and $\mathbf{b}'_{2,2}$ as examples, $\mathbf{W}'_{2,2} = \mathbf{W}'_{2,2}\mathbf{W}'_{1,1}$, where $\mathbf{W}'_{1,1}$ is an input-dependent parameter from MHA, and $\mathbf{W}'_{2,2}$ is a defined parameter, becoming fixed post-training. Thus, the degrees of freedom for $\mathbf{W}'_{2,2}$ is imparted through a linear transformation with the MHA parameter $\mathbf{W}'_{1,1}$. Additionally, $\mathbf{b}'_{2,2} = (\mathbf{W}'_{2,2}\mathbf{b}'_{1,3} + \mathbf{b}'_{2,2}) + \mathbf{W}'_{2,2}\mathbf{W}'_{1,3}\sigma(\mathbf{W}'_{1,2}\mathbf{x}'_{0} + \mathbf{b}'_{1,2})$. Hence, $\mathbf{b}'_{2,2}$  can be understood as being approximated by a two-layer DUAT, showcasing stronger dynamic transformation capabilities based on inputs. Due to DUAT's powerful capacity, $\mathbf{b}'_{2,2}$ can dynamically fit specific functions based on inputs.

In a single-layer Transformer, as seen in Eq. \eqref{eq:TF-app-1}, $\mathbf{b}'_{1,3}$ and $\mathbf{b}'_{1,2}$ are defined parameters, so the number of bias terms approximated using UAT is zero. For a two-layer Transformer, the number of bias terms approximated using DUAT is one, specifically  $\mathbf{b}'_{2,2}$ with the DUAT layers approximating this bias terms being 1. According to Eq.\eqref{eq:TF-app}, in a Transformer with $i+1$ layers, $\mathbf{b}_{i+1,1}$ is approximated by a single-layer UAT, while $\mathbf{b}'_{j,2}$ is approximated by a UAT of $j$ layers for $j > 1$. It is easy to deduce that for a three-layer Transformer, the number of bias terms approximated using UAT is three, with the UAT layers being 1, 2, and 3. For a six-layer Transformer, the number of bias terms approximated using DUAT is six, with the DUAT layers being 1, 2, 3, 4, 5, and 6. The  DUAT layers of bias terms for Para-F-1-24, Para-F-6-6, Para-F-2-24, Para-F-3-24, and Para-F-6-24 are shown in Table \ref{tab:UAT_B}.

\begin{table*}[htbp!]
\centering
\begin{tabular}{lccclc} 
\toprule
                                      & Para-F-1-24 & Para-F-6-6  & Para-F-2-24 & Para-F-3-24                   & Para-F-6-24                                                \\ 
\hline
\multicolumn{1}{c}{UAT Layers} & (0)*24        & (1+2+3+4+5+6)*1 & (1+2)*12    & {(1+2+3)*8} & (1+2+3+4+5+6)*4 \\
\bottomrule
\end{tabular}
\caption{The DUAT layer count corresponding to the bias terms in Para-F-1-24, Para-F-6-6, Para-F-2-24, Para-F-3-24, and Para-F-6-24 is calculated by summing the UAT layers associated with each bias term within a serial Transformer. Since each serial Transformer has multiple bias terms, with each term corresponding to different UAT layer counts, we represent the UAT layers by summing the UAT layers for each bias term. For example, "1+2+3" indicates that three bias terms are approximated using UAT, with 1, 2, and 3 layers respectively. Additionally, since multiple serial Transformers are used in parallel, we multiply the sum by the number of parallel Transformers.}
\label{tab:UAT_B}
\end{table*}

In summary, more layers mean more degrees of freedom in bias terms, leading to stronger model fitting capabilities. To achieve performance comparable to that of deep serial networks while ensuring model fitting ability and fast inference, parallel networks need to strike a balance between network depth and layer count. Parallel networks are designed for extremely large models. For instance, if a serial network with 1 million layers can be replaced by a parallel network with a depth of 10 and 100 million layers, the speedup factor can reach up to 100,000 times. Sure, we could improve the design of the Para-former to achieve speedup computing with fewer layers.

\end{document}